%% file: CCTN_arxiv.tex
\documentclass[10pt,twocolumn,letterpaper]{article}

\usepackage{cvpr}
\usepackage{times}
\usepackage{epsfig}
\usepackage{graphicx}
\usepackage{amsmath}
\usepackage{amssymb}

\usepackage{multirow}
\usepackage{subfigure}
\usepackage{algorithm}
\usepackage{algorithmic}
\usepackage{subfigure}

\usepackage[pagebackref=true,breaklinks=true,letterpaper=true,colorlinks,bookmarks=false]{hyperref}

\cvprfinalcopy 

\def\cvprPaperID{2429} 

\ifcvprfinal\fi
\begin{document}

\title{Accurate Text Localization in Natural Image with Cascaded \\ Convolutional Text Network}

\newcommand*\samethanks[1][\value{footnote}]{\footnotemark[#1]}
\author{Tong He$^{1,}$ $^2$, Weilin Huang$^{1,}$ $^3$, Yu Qiao$^1,$ $^3$, and Jian Yao$^{2}$
       \\
       $^1$
       Shenzhen Institutes of Advanced Technology, Chinese Academy of Sciences, Shenzhen, China
       \\
       $^2$School of Remote Sensing and Information Engineering, Wuhan University, Wuhan, China
       \\
       $^3$Department of Information Engineering, The Chinese University of Hong Kong, Shatin, Hongkong\\
       {\tt\small \{tong.he,wl.huang,yu.qiao\}@siat.ac.cn;jian.yao@whu.edu.cn}
}




\maketitle

\input{abstract}

\input{introduction}
\input{relatedwork}

\input{methods}

\input{experiment}

\input{conclusion}

{\small
\bibliographystyle{ieee}
\bibliography{egbib2}
}

\end{document}

%% file: abstract.tex
\begin{abstract}
We introduce a new top-down pipeline for scene text detection. We propose a novel Cascaded  Convolutional Text Network (CCTN) that joints two customized convolutional networks for coarse-to-fine text localization.
The CCTN fast detects text regions roughly  from a  low-resolution image, and then accurately localizes text lines from each enlarged region. 
We cast  previous character based detection into direct text region estimation, avoiding multiple bottom-up post-processing steps. It exhibits surprising robustness and discriminative power by considering whole text region as  detection object which provides strong semantic information. 
We customize convolutional network by developing rectangle convolutions and multiple in-network fusions. This enables it to handle multi-shape and multi-scale text efficiently. Furthermore, the CCTN is computationally efficient by sharing convolutional computations, and  high-level property allows it to be invariant to various languages and multiple orientations. It achieves 0.84 and 0.86 F-measures on the
ICDAR 2011 and ICDAR 2013, delivering substantial improvements over state-of-the-art results \cite{Tian2015,Busta2015}.

%
\end{abstract}

%% file: introduction.tex
\section{Introduction}
\label{Sec:Introduction}

 \begin{figure}
\centering
\subfigure{\includegraphics[height=8.5cm,width=8.5cm]{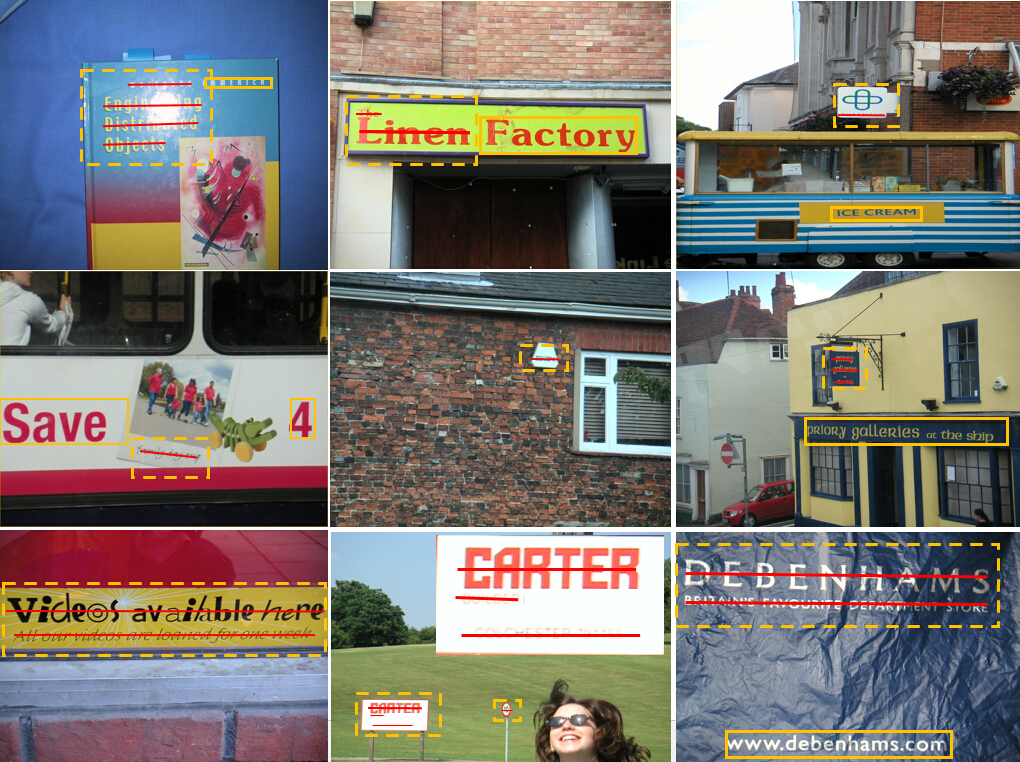}}
\vskip -0.1cm
\caption{Two-step coarse-to-fine text localization results by the proposed Cascaded Convolutional Text Network (CCTN). A coarse text network detects text regions (which may include multiple or single text lines) from an image, while a fine text network further refines the detected regions  by accurately localizing each individual text line.  The ORANGE bounding box indicates a detected region by the coarse text network. We have two options for each text region: (i)  directly output the bounding box as a final detection (solid ORANGE); (ii) refine the detected region by the fine text network (dashed ORANGE), and generate an accurate location for each text line (RED solid central line). The refined regions may include multiple text lines or an ambiguous text line (e.g., very small-scale text).}
\label{fig:main}
\end{figure}

Text detection and recognition in natural images have recently gained increasing attention from computer
vision community, as substantiated by recent work~\cite{Jaderberg2015,Huang2014,Ye2015,Jaderberg2014,He2016P,Busta2015,Yin2015,Tian2015,Zhang2015,Yin2014}.
This paper focuses on text detection sub task. 
Though tremendous efforts have been devoted to improving its performance, accurately locating text in unconstrained environments is still extremely challenging, due to  significant diversity of text patterns and highly complicated background.
For example,  text can be in  very small size, low quality, or low contrast, and even regular ones can be distorted considerably by numerous real-world affects, such as perspective transform, strong lighting, large-scale occlusion, or blurring. 
These pose fundamental challenges of this task where correct character detection is difficult, and multiple post-processing steps that group character candidates into text lines are highly complicated and unreliable.

%

Most existing scene text detection methods are built on bottom-up strategy that sequentially processes: \emph{stroke or character candidate detection}, \emph{filtering}, \emph{text line construction} and \emph{classification}~\cite{Jaderberg2014,Yin2015,Huang2014,Yin2014,Huang2013,Neumann2014}. These approaches commonly suffer from a number of limitations. First, detecting strokes or characters by exploring low-level image cues is not robust, e.g., by using the widely-used SWT \cite{Epshtein2010} or MSERs \cite{Matas2004} detector. Second, it is easy to generate a large amount of non-text candidates, which can be many orders of magnitude more than the true text candidates. This makes it extremely challenging to filter out these non-text false detections robustly by using a character level classifier. Third,  grouping the retained character candidates into text lines is complicated. It often explores a number of low-level heuristic properties and geometric information, and also requires manually setting a number of low-level grouping rules. Fourth, as indicated in \cite{Tian2015}, the bottom-up strategy is not reliable, where error in each step can be accumulated sequentially.

These limitations severely harm the performance of current systems, and recent efforts have been given to tackle one or some of them. Building on recent advances of deep learning models for image representation, Jaderberg \emph{et al.}~\cite{Jaderberg2014} and Wang \emph{et al.}~\cite{Wang2012} designed their Convolutional Neural Networks (CNN) to detect character information in a sliding window fashion. The CNN models were also applied for filtering out non-character candidates by Huang \emph{et al.} \cite{Huang2014}, or for assigning character confident scores in \cite{Tian2015}. These approaches achieved the state-of-the-art performance by leveraging strong representation-capability of the CNN. 
%
However, these approaches commonly employ the CNN models for character-level classification, which is neither robust nor discriminative. It is more principled to jointly identify a group of text strings (e.g. a text region) where local neighbouring text is greatly helpful to make a reliable decision.
Therefore, current text models did not fully explore excellent potential of the CNN as exhibited in generic object detection or segmentation. 

In this  work,  we fill this gap by introducing CNN to direct text region estimation. 
Conventional CNN architecture includes a number of stacked convolutional layers followed by several fully-connected (FC) layers. The FC representation, which discards spatial information, is particularly efficient for classification task, but is not effective for localization task. 
Long \emph{et al.} replaced the  FC connections with 1$\times$1 convolutions, and achieved fully convolutional networks for semantic segmentation \cite{Long2015}. 
This inspires us to utilize the fully convolutional properties which preserve coarse spatial information of the image. 
In the convolutional architecture, pooling operation can reduce computational complexity, and also introduce invariance to local transformations. However, these advantages come at the price of reduced spatial accuracy, which is of particular importance for accurate text localization. 
To overcome these limitations, we propose a two-stage coarse-to-fine pipeline that tailors general convolutional networks towards our problem.



In this work, we tackle problem of text localization from an alternative perspective. We propose a Cascaded  Convolutional Text Network (CCTN) for direct text region estimation. We develop an efficient top-to-down pipeline that localize text in a coarse-to-fine manner (see Fig. \ref{fig:main}),  departing from previous approaches building on character based detection. It makes the following major contributions.

First, this is the first attempt to cast previous character based detection approaches into direct text region estimation. It provides surprising robustness and discriminative power by considering whole text region as detection object. Our approach dose not include any post-processing, providing a significant simplification of existing methods.

Second, we customize general convolutional networks towards our text task. We design rectangle convolutions and multiple in-network
fusions to handle multi-shape and multi-scale text lines. This allows our network to work reliably on a single-scale  input image, compared to the 24 scales used in \cite{Zhang2015}. Besides, our fully convolutional design further reduces  computational complexity.

Third, we develop a coarse-to-fine strategy which improves localization accuracy lost in pooling and up-sampling operations. We design a fine text network that estimates locations of central line and text line area. It separates text regions into text lines accurately, and also reduces false region detections, as shown in Fig.~\ref{fig:pipeline}.

Fourth, our CCTN computes high-level deep features, giving excellent generalization ability. It achieves promising results on the standard ICDAR  2011 and 2013 datasets, with F-measure of 0.84 and 0.86, outperforming previous methods substantially. It also obtains excellent performance on the multilingual and multi-orientation MSRA-TD500 \cite{Yao2012}, by just training on English text.

%% file: relatedwork.tex
\section{Related Work}

There are two groups of methods for text detection and localization in natural images: connected-component and sliding-window based approaches. The connected-component approaches detect text information at pixel-level by using a fast low-level detector, and then group the detected pixels into text component candidates. While the sliding-window methods scan the image densely by using a multi-scale sub-window with a pre-designed classifier.

 \begin{figure*}[ht]
\begin{center}
\subfigure[Image]{\includegraphics[height=6cm, width=2.8cm]{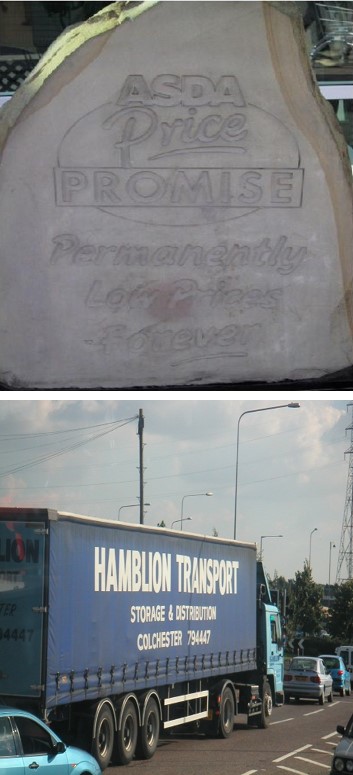}}
\subfigure[TR-Heatmap]{\includegraphics[height=6cm, width=2.8cm]{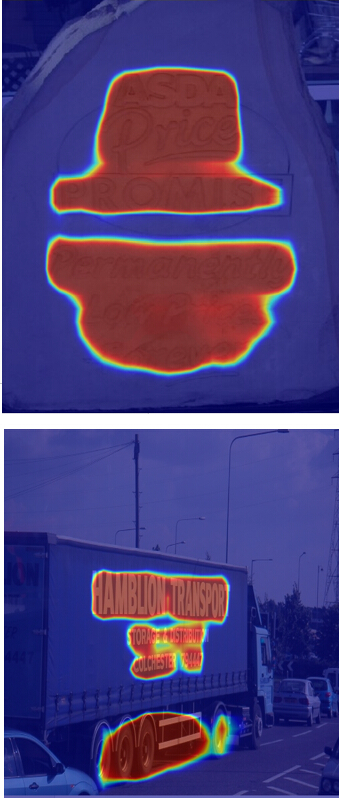}}
\subfigure[Text Region]{\includegraphics[height=6cm, width=2.8cm]{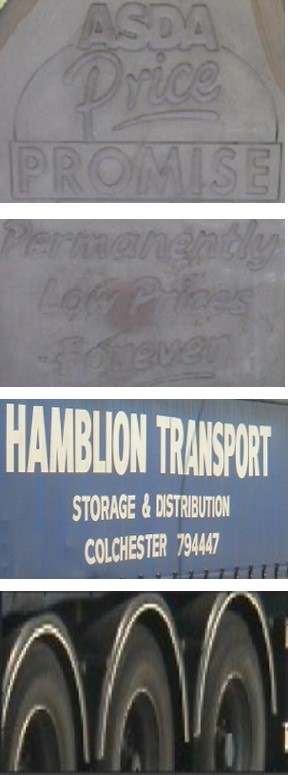}}
\subfigure[CL-Heatmap]{\includegraphics[height=6cm, width=2.8cm]{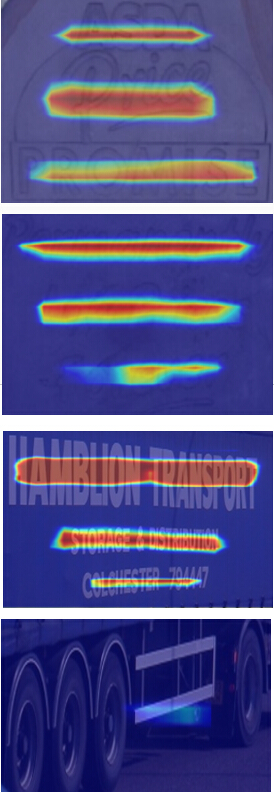}}
\subfigure[TX-Heatmap]{\includegraphics[height=6cm, width=2.8cm]{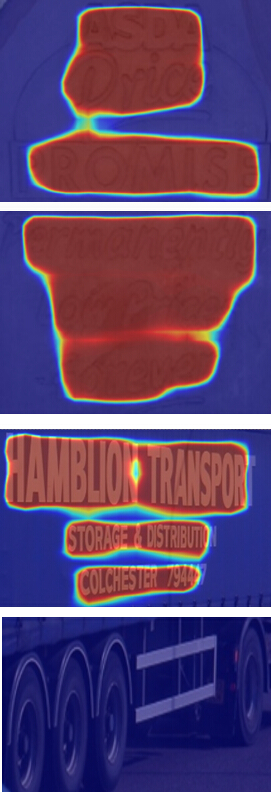}}
\subfigure[Text Line]{\includegraphics[height=6cm, width=2.8cm]{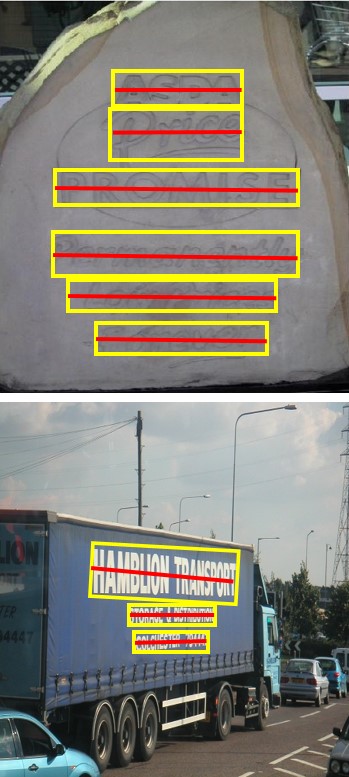}}
\end{center}
\vskip -0.4cm
   \caption{Pipeline of the Cascaded Convolutional Text Network (CCTN), which includes a coarse text network and a fine text network. (a) The input image (resized to 500$\times$500); (b) The coarse text network detects text regions roughly by generating a text region heat-map (TR-Heatmap); (c) The detected regions are cropped out and enlarged to 500$\times$500; (d-e) The fine text network refines them, and generate a central line area heat-map (CT-Heatmap) and a text line area heat-map (TL-Heatmap) for each region; (f) Final detection results.}
\label{fig:pipeline}
\end{figure*}

The connected-component approaches have recently achieved promising performance ~\cite{Yin2014,Huang2014,Yin2015,Huang2013,Kang2014,Li2014,Yao2012,Epshtein2010}. Among them, the Stroke Width Transform (SWT)~\cite{Epshtein2010} and Maximally Stable Extremal Regions (MSERs)~\cite{Matas2004} are two representative low-level methods for detecting text component candidates. The MSERs algorithm is powerful to detect challenging text patterns,  resulting in a good recall in character detection. This leads to a number of high-performance systems ~\cite{Yin2014,Huang2014,Kang2014}. Recent efforts on developing the low-level text detector include Stroke Feature Transform (SFT) ~\cite{Huang2013}, Characterness \cite{Li2014}, and  EdgeBox ~\cite{Zitnick2014,Jaderberg2015}. Generally, the connected-component approaches based on these low-level detectors have a great advantage in speed by tracking image pixels in one pass computation. However, these detectors often generate a large amount of non-text components due to their low-level nature, raising 
 main difficulties on filtering these non-text components, and grouping components into text lines. 
Consequently, they often require a number of bottom-up post-processing steps to reach a good performance. In fact, developing these post-processing methods is difficult itself. Previous methods generally develop a number of heuristic rules by exploring various low-level image properties,  making the whole systems highly complicated and unreliable.
 
  \begin{figure*}
\begin{center}
\subfigure[]{\includegraphics[height=4cm, width=14cm]{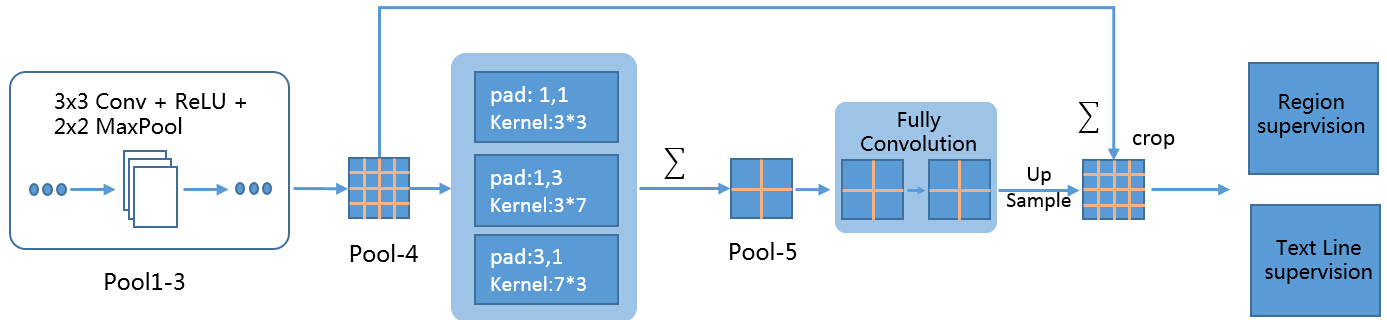}}\hspace*{0.9em}
\subfigure[]{\includegraphics[height=4cm, width=2cm]{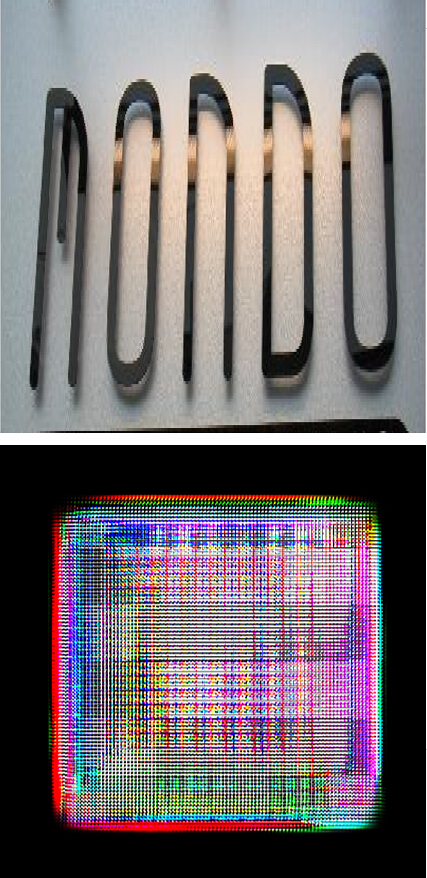}}
\end{center}
\vskip -0.4cm
   \caption{ (a) Structure of the Convolutional Text Network, which is built on 16-layer VggNet \cite{Simonyan2015}. (b) An resized 500$\times$500 input image, and the actual receptive filed of new $Pool$-5, which is computed as the response area in the input image by propagating the error of a single neuron in the new $Pool$-5.}
\label{fig:textNets}
\end{figure*}

The sliding-window based methods have  been developed rapidly~\cite{Jaderberg2014,Tian2015,Zhang2015,Wang2011,Wang2010,Wang2012,Chen2004}. One obvious advantage is that it computes global feature from a scanned window, so that the feature is invariant to various low-level distortions or transformations. However, they have a number of drawbacks. (i) A major challenge is the high computational cost. It requires mutli-sclae windows to handle various font sizes of text, resulting in a large number of the scanning windows. (ii) Designing a discriminative feature and training a powerful text/non-text classifier are difficult. (iii) Existing approaches are mostly built on character level detection, which is not robut and unreliable. They also require a number of bottom-up steps to refine detection results, such as identifying and grouping character candidates into text lines, increasing complexity of the systems considerably \cite{Jaderberg2014,Tian2015,Wang2012,Chen2004}.

Current leading performance are achieved by incorporating CNN models to improve one or multiple components of the systems \cite{Jaderberg2015,Huang2014, He2016,Zhang2015,Tian2015,Jaderberg2014,Wang2012}.  CNN is powerful to compute a high-level image feature, which is particularly efficient for classification task. Therefore, most of current approaches apply a CNN model for character/non-character classification. For example,  Huang \emph{et al}. \cite{Huang2014} and He \textit{et al.} \cite{He2016} both exploited a CNN classifier to filter out non-character MSERs components, while sliding-window approaches built on a CNN character classifier were developed in \cite{Jaderberg2014,Wang2012}. Although these CNN models have largely improved the performance of previous manually-designed features,  the potential of CNN is not fully explored by these methods. Applying a  CNN model for character classification is unreliable and inefficient. We customize general CNN  towards our text localization task, resulting in a much higher accuracy.

Recently, Tian \emph{et al}. proposed TextFlow, which simplifies multiple post-processing steps by utilizing a minimum cost flow network \cite{Tian2015}.  They applied cascade boosting algorithm with a number of manually-designed features for detecting character candidates. Then a trained CNN was used to assign a score to each character candidate. This approach is still built on character-level detection and classification. Alternatively, our method discriminates text and background information using a whole text region, which is more robust and discriminative. Our work is also related to Zhang \emph{et al}.'s work where a text line is discriminated with a group of neighboring characters  \cite{Zhang2015}. They designed a number of low-level symmetry features for detecting text line components, and also proposed multiple post-processing approaches with a number of heuristic rules and low-level properties to group the detected components into text lines, and post refine final results. Besides, the symmetry features are difficult to be generalized  to oriented and multi-scale text lines. Here, we provide a more efficient approach that generalizes better.

%% file: methods.tex
\section{Cascaded Convolutional Text Network}


In this section, we present details of the proposed Cascaded  Convolutional Text Networks (CCTN). It includes a coarse text network  for rough text region detection from the whole image, and a fine text network  used to accurately estimate a central line and a fine-grained region for each text line within an enlarged region. Our pipeline is presented in Fig. \ref{fig:pipeline}. The coarse network directly outputs a per-pixel heat-map, densely indicating the location and probability of text information. The fine network generates two such heat-maps for each cropped region. One heat-map presents  the location of  central line for each text line, and the other one gives separated areas of the text lines. In this work we show effectiveness of exploring convolutional networks to directly output accurate text regions, departing from previous approaches that apply a CNN classifier in a sliding window fashion for text detection, which require a number of complicated post-processing steps.

\subsection{Coarse Text Network}
\begin{figure}
\begin{center}
\includegraphics[width=8cm]{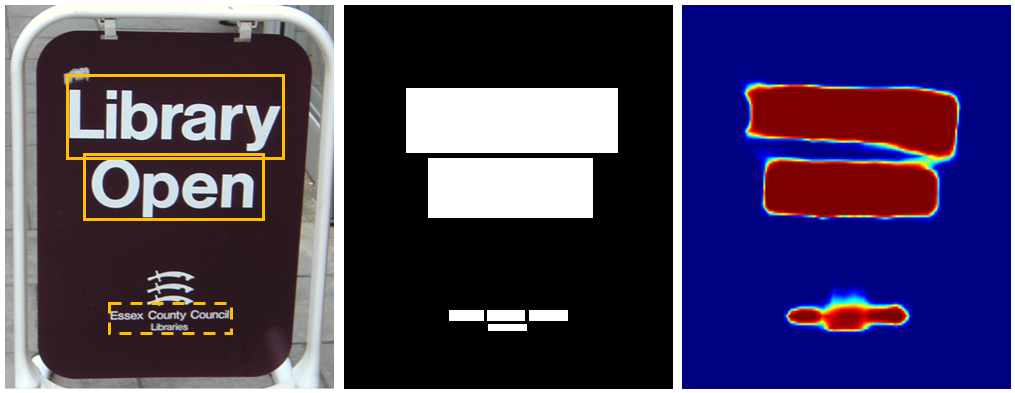}
\end{center}
\vskip -0.4cm
   \caption{Text region mask (middle) and heat-map (right).}
\label{fig:TextRegionLabel}
\end{figure}

\begin{figure*}
\begin{center}
\includegraphics[height=2cm,width=2.4cm]{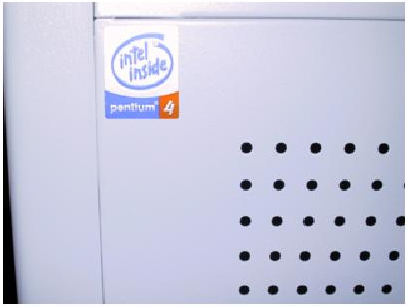}
\includegraphics[height=2cm,width=2.4cm]{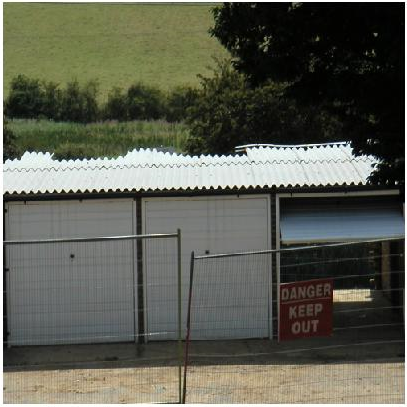}
\includegraphics[height=2cm,width=2.4cm]{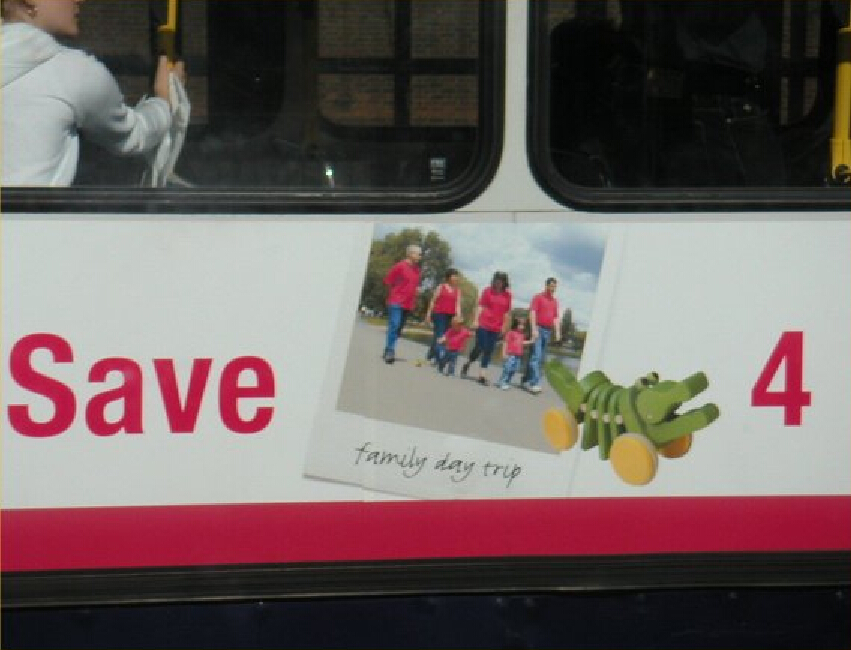}
\includegraphics[height=2cm,width=2.4cm]{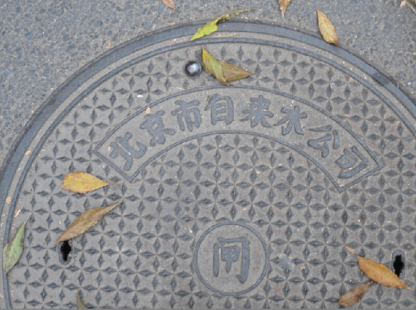}
\includegraphics[height=2cm,width=2.4cm]{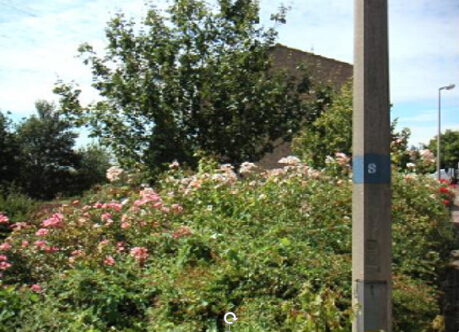}
\includegraphics[height=2cm,width=2.4cm]{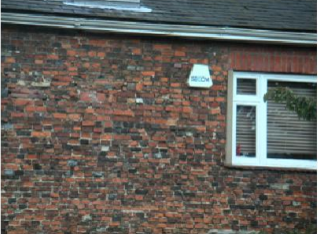}
\includegraphics[height=2cm,width=2.4cm]{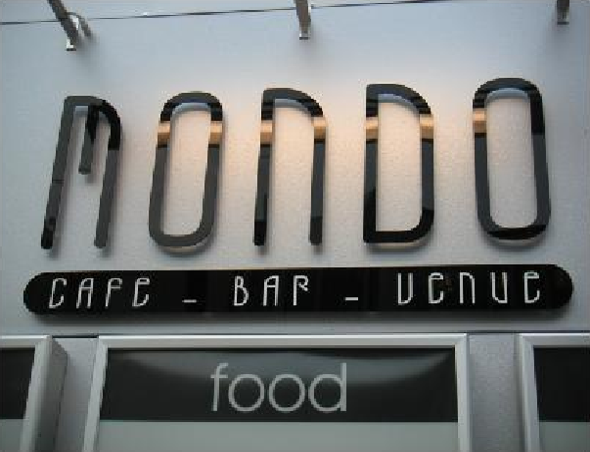}\\
\vskip 0.05cm

\includegraphics[height=2cm,width=2.4cm]{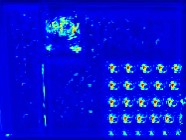}
\includegraphics[height=2cm,width=2.4cm]{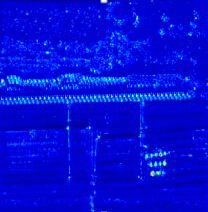}
\includegraphics[height=2cm,width=2.4cm]{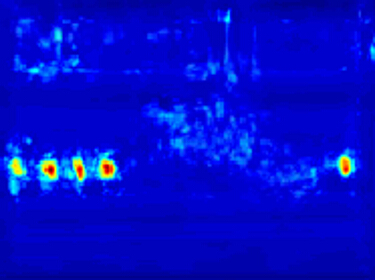}
\includegraphics[height=2cm,width=2.4cm]{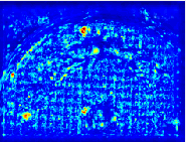}
\includegraphics[height=2cm,width=2.4cm]{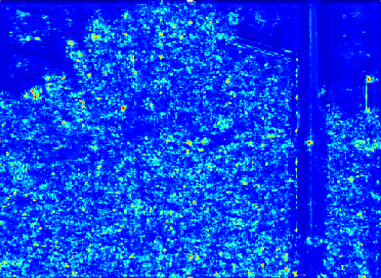}
\includegraphics[height=2cm,width=2.4cm]{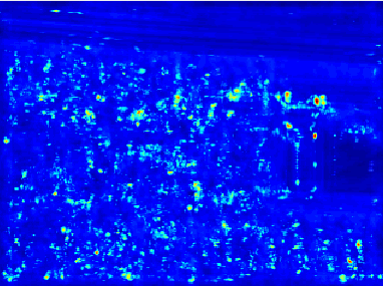}
\includegraphics[height=2cm,width=2.4cm]{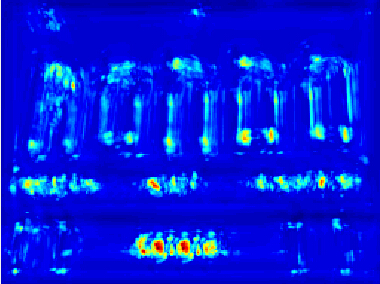}\\
\vskip 0.05cm

\includegraphics[height=2cm,width=2.4cm]{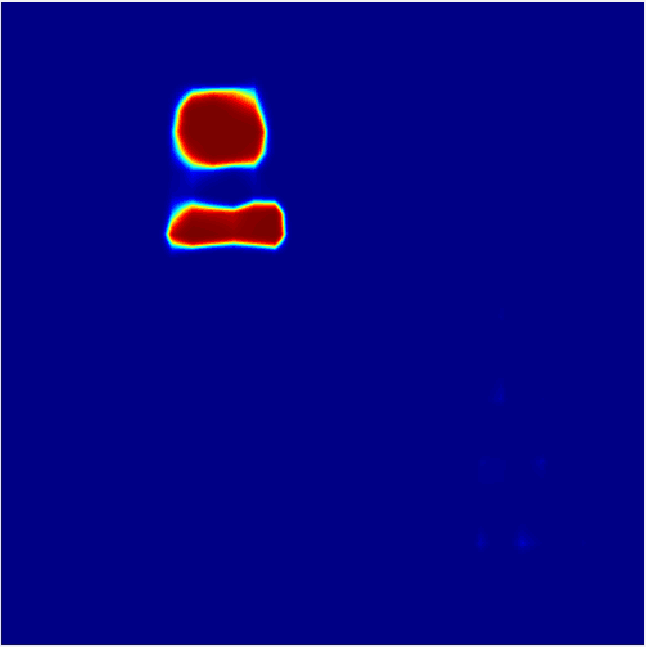}
\includegraphics[height=2cm,width=2.4cm]{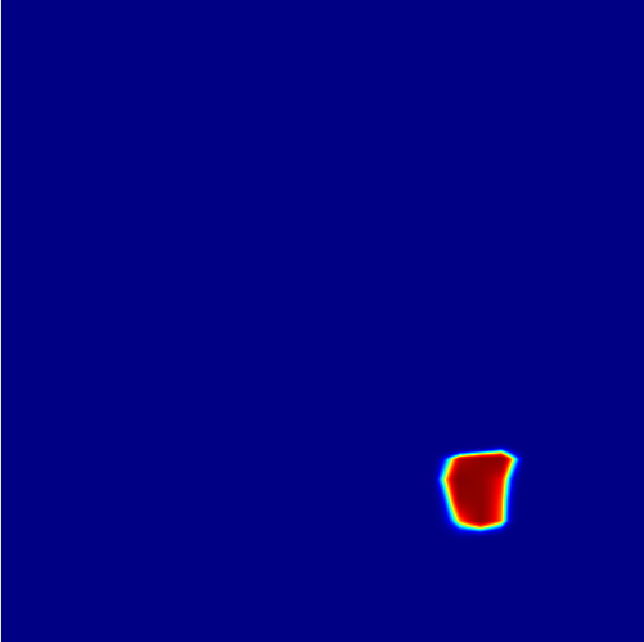}
\includegraphics[height=2cm,width=2.4cm]{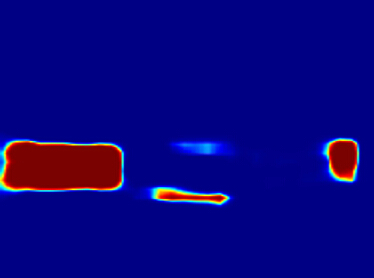}
\includegraphics[height=2cm,width=2.4cm]{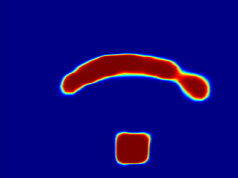}
\includegraphics[height=2cm,width=2.4cm]{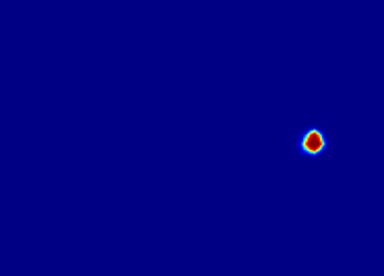}
\includegraphics[height=2cm,width=2.4cm]{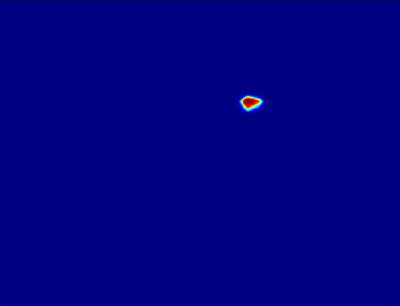}
\includegraphics[height=2cm,width=2.4cm]{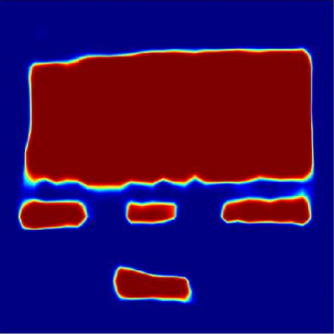}
\end{center}
\vskip -0.4cm
   \caption{Comparisons of our heat-maps (Bottom) with those generated by CNN based character detector with a sliding-window (Middle). The CNN character/non-character classifier is provided by the authors of \cite{Jaderberg2014}. We select the best map of six scales for comparisons, while our coarse text network is just run on the single-scale image. Obviously, the results suggest that  our method is strongly robust against cluttered background, and is also powerful to identify ambiguous text. It works reliably on both small-scale and large-scale text.}
\label{fig:heatmap_ctn}
\end{figure*}

The goal of our coarse network is to fast locate coarse text regions through a whole low-resolution image, with powerful discriminative ability and strong robustness against complicated background information. Most previous CNN based approaches for text localization train a character level CNN classifier and apply it for scanning an image densely with multi-scale sliding windows. They generate a corresponding heat-map that indicates the probabilities and locations of the text \cite{Jaderberg2014, Wang2012}. These approaches are built on character level detection, which suffers from two limitations. First, it is extremely difficult to robustly distinguish isolated characters from complicated background outliers, imposing a large number of false detections, as shown in Fig. \ref{fig:heatmap_ctn}. Second, it requires a number of complicated post-processing steps to remove the false detections, and to bottom up the detected characters into text lines,  making their systems highly complicated. 
Third,  it is difficult to handle large diversity of the font sizes efficiently, and exploring multi-scale windows largely increases computational demands. 
To circumvent these problems, we present a new convolutional architecture that directly outputs a reliable text region heat-map. We cast conventional character classifier CNN into our text region estimation network. The output text heat-maps are shown in Fig. \ref{fig:pipeline}.

Our convolutional text network is built on the widely-used 16-layer VggNet architecture which has 16 convolutional layers separated by 5 pooling layers \cite{Simonyan2015}. Inspired by design of fully convolutional networks (FCN) for sematic segmentation \cite{Long2015}, we apply $1\times1$ convolutional layers for replacing the FC layers of original VggNet. This makes our network fully convolutional and capable of preserving rough spatial information. The structure of our convolutional text network is presented in Fig. \ref{fig:textNets}, where we highlight our customized designs that make it adaptive to our text detection task.

\textbf{Text Region Supervision}. Our network is encouraged to directly output a text region estimation, e.g., a heat-map indicating the probabilities of text in all pixel locations. We consider all pixels within a text line area as text pixels. They include those pixels which are not exactly located on text strokes, but are inside the bounding box area of a text line. Because respective fields (RFs) of these pixels certainly include meaningful local text information around them. Thus we use a pixel-wise text region mask as our supervision information for region  estimation. This information is obtained cheaply by just labeling all pixels within a text line bounding box. A exemplar region mask is presented in Fig. \ref{fig:TextRegionLabel}. The network is trained with pixel-wise softmax loss. This simple labeling scheme allows our detector to jointly consider local neighboring information, making it surprisingly advanced over low-level pixel based detectors, such as SWT and MSERs. It also significantly superior to those character based CNN detectors used in \cite{Jaderberg2014,Wang2012}, as compared in Fig. \ref{fig:heatmap_ctn}.


\textbf{Text Rectangle Kernels}.  We consider a text line as our detection object. We found that appearance of the text lines has obvious characteristics. Their shapes are commonly in square (e.g., a word just including one or two characters), horizontal rectangle or vertical rectangle (e.g., oriented text). To handle such text-specific properties, we design three parallel convolutional layers with various kernel shapes to replace original three convolutional layers between the $Pool$-4 and $Pool$-5 layers. The three layers in original VggNet are sequentially connected by using a same $3\times3$ convolutional kernel. As shown in Fig. \ref{fig:textNets}, sizes of the three kernels are set to $3\times3$, $3\times7$ and $7\times3$ respectively in our text network. Differing from the original sequential architecture, we  parallelize the three layers, and design different padding sizes for them, which allow them to output three equally-sized maps. This design allows the activations in three layers to have their RFs in a wide range of shapes and aspect ratios, making them capable of detecting text in various shapes efficiently.


\textbf{Multi-Layer Fusion}. We design a two-step fusion strategy. In the first step, we fuse  output maps of  the designed three parallel layers into a single layer, by using element-wise summarization. Then the fused maps are further max-pooled with a $2\times2$ kernel to procure new $Pool$-5 maps. This operation enlarges the RFs by $2\times2$ in this layer. To achieve multi-scale capability, a straightforward approach is to combine current feature maps with the output maps of previous layers, which capture more local fine-scale features by using smaller RFs. In the second step, to have a equal map size as the previous layer (e.g., the $Pool$-4 maps),  the $Pool$-5 maps are first $2\times$-upsampled (after passing two 1$\times$1 fully convolutional layers), and then are element-wisely combined with the $Pool$-4 maps. Notice that this up-sampling operation does not change the sizes of RFs in $Pool$-5. 
In practice, in all our implementations we resize each input image to 500$\times$500. In our architecture, the actual RFs in the new $Pool$-5 is 403$\times$403, which is able to cover most area of input image, as shown in Fig. \ref{fig:textNets} (b). The actual RFs is computed as the response area in the input image by propagating the error of a single neuron in the new $Pool$-5.
After the two-step combination, the feature maps in the new $Pool$-5 are able to capture both multi-shape (square, horizontal or vertical rectangle) and multi-scale text lines. \textit{Therefore, our customized text network is strongly robust against cluttered background, and is powerful to identify ambiguous text. It works reliably on both small-scale and large-scale text with a single-scale input, successfully avoiding expensive computational cost raised by exploring multi-scale sliding-windows. }
Examples and comparisons are presented in Fig. \ref{fig:heatmap_ctn}.




\textbf{Text Region/ Text Line Extraction}. We observed that some large-scale text lines which do not have closed neighbouring lines can be localized  accurately in the coarse heat-map. However, the estimated areas of multiple closing text lines or small-scale ones are easily merged and generally inaccurate. This may due to the max-pooling operation and low-resolution image used, as shown in Fig. \ref{fig:TextRegionLabel}
 (c). For those accurately detected text lines, we directly extract their bounding boxes from the coarse heat-map. The ambiguous regions are cropped out and further refined by the proposed fine network, as described in Fig. \ref{fig:main}. We design a simple rule that extracts the text lines or text regions as follow:
(1) We binary the heat-map with a threshold of 0.3. 
(2) We compute area ratio and borderline ratio.
(3) We directly output a text line bounding box if the area ratio $>0.7$ and borderline ratio $>5$, while leaving the others (text regions) for a further refinement.   
(4) We crop the remained text regions. A text region is cropped by enlarging it in square, with a side length of 1.2 $\times$ longer side.

 Our text network directly outputs a pixel-wise heat-map which labels the input image densely. It essentially works in a sliding window fashion. So we compare our results to those of Jaderberg \emph{et al}.'s method \cite{Jaderberg2014} in Fig. \ref{fig:heatmap_ctn}. They also used the sliding-window approach with a character/non-character CNN classifier for generating the text heat-maps. As can been seen, our heat-maps are surprisingly better than those generated by character based CNN detector. Our convolutional text network is very robust to complicated background outliers, and strongly discriminate ambiguous text regions. It is able to handle multi-scale and multi-shape text  reliably in a single scale. The excellent performance are mainly ascribed to our region based detection strategy and multiple in-network fusions. They allow our detector to evaluate each single pixel reliably by inherently incorporating multi-scale and multi-shape image content. Furthermore, our method does not require any additional post-processing steps to bottom up character candidates into text lines. This grouping step is difficult when it is implemented on the character heat-maps, where many false detections exist, as shown in Fig. \ref{fig:heatmap_ctn}. 
 
 Despite with these appealing properties, the coarse network can not provide accurate bounding boxes for all text lines. The localization accuracy is significantly reduced by multiple pooling and sub-sampling operations, so that a further refinement is necessary.

\subsection{Fine Text Network}

Our coarse network is able to detect text information in rough regions reliably with few false detection. Our ultimate goal is to find accurate text area in text line or word level. Although the detected coarse regions can be used to find  final detections of some isolate large-scale text lines,  localization accuracies on small-scale text fonts or multiple closed text lines are significantly reduced by using the low-resolution image (e.g. 500$\times$500 in our experiments) and multiple pooling operations. While these strategies are indeed beneficial for speed and robustness  by allowing larger RFs to scan image content, it is principled to develop a fine detection network to refine the coarse detection results (e.g., the cropped text regions). The goal of our fine network is to correctly separate all isolated text lines within a text region, and also remove false detections in the coarse detections (despite the number is  small).
\begin{figure}
\begin{center}
\includegraphics[height=1cm,width=2.6cm]{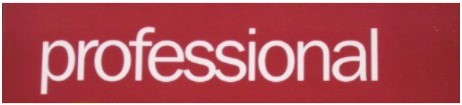}
\includegraphics[height=1cm,width=2.6cm]{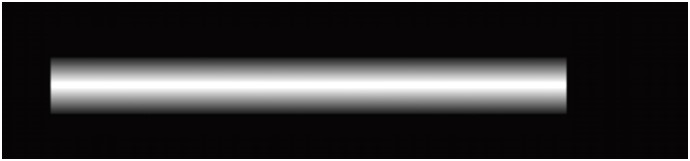}
\includegraphics[height=1cm,width=2.6cm]{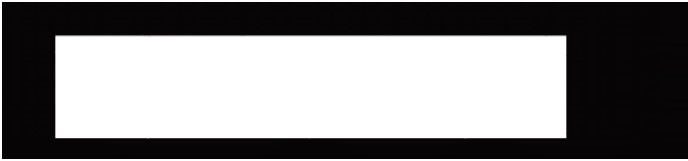}
\end{center}
\vskip -0.3cm
   \caption{Central line area (middle) and text line area (right).}
\label{fig:TextLineLabel}
\end{figure}

\begin{figure*}
\begin{center}
\includegraphics[width=17cm]{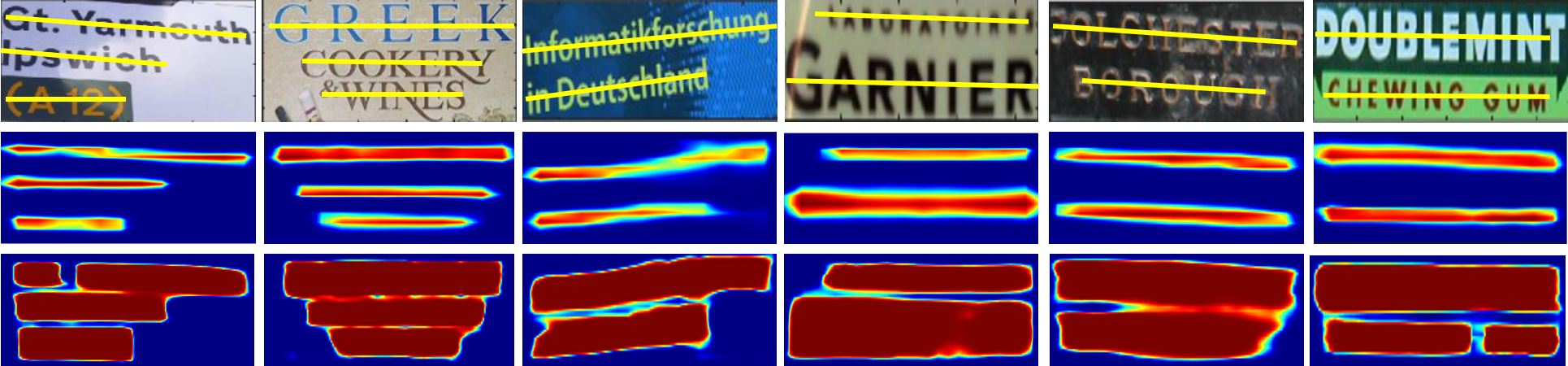}
\end{center}
\vskip -0.3cm
   \caption{Results of the fine text network. \textbf{Top Row}: image and central line detection; \textbf{Middle Row}: central line area heat-map (CT-Heatmap); \textbf{Bottom Row}: text line area heat-map (TL-Heatmap).}
\label{fig:FTN_examples}
\end{figure*}

We look for a discriminative property that can separate multiple closed text lines correctly. A straightforward approach is to use text area within each text line bounding box, which is ideally separable. In practice, it may not work well in two situations. First, the text line areas are easily overlapped between two closed yet oriented text lines. Second, even the areas of two text lines are  clearly separable in the original image, they may be confused in the estimated heat-map by multiple pooling and up-sampling operations, as shown in Fig. \ref{fig:TextRegionLabel}. Thus purely estimating the area of a text line is not reliable, and  a more fine-grained text line localization is required. Therefore, we enlarge each detected text region to refine text line locations by using a fine text network. The fine text network has a same architecture  as the coarse text network (in Fig. \ref{fig:textNets} (a)), but with a different output layer which we will describe bellow. Since text lines in an enlarged region are generally in large scale,  e.g., covering most spatial area of a text region, the RFs of the network should be able to cover full area of an input region. Thus we resize each text region into a fix size of 500$\times$500 with 50-pixel (zero values) padding on each side, so that full image content is constrained into the central area of 400$\times$400, which is fully covered by a  RFs of 403$\times$403. 

\textbf{Text Line Supervision}. Motivated from the symmetry-based detector in \cite{Zhang2015}, we found that a central line is more reliable to define an unique text line, providing a better choice for our design. We aim to locate each text line separately by using a bounding box, so that we also need to measure the height of a text line if we know the location of its central line. With these considerations, we design a fine text network that is able to jointly estimate both \emph{central line area} and \emph{text line area}, as shown in Fig. \ref{fig:TextLineLabel}. The \emph{central line area} is defined by using a Gaussian distribution with its maximum value (e.g., 1) in the middle of a bounding box, which is decreased to 0 in a radius of $0.25\times H$, where $H$ is the height of bounding box. We use half height of the bounding box as the height of central line area. This design allows it to include both central line location and height information of a text line, leading to a better separation between closed text lines than using a full height. The \emph{text line area} is all pixels within its bounding box. It is used to measure the height of a text line. Therefore, the fine text network is trained with both supervision masks by jointly computing per-pixel cross-entropy and softmax loss; and it should be able to estimate both the central line area and text line area heat-maps given an cropped region. The output heat-maps are presented in Fig. \ref{fig:pipeline} (d) and (e).

\textbf{Text Line Extraction}
For accurately locating a detected text line, we compute the central line and height of  bounding box by using both central line area heat-map (CL-Heatmap) and text line area heat-map (TL-Heatmap).
(1) We binary both heat-maps by using a threshold of 0.5.
(2) Then we compute a minimum area rectangle (MAR) from the binarized CL-Heatmap, and compute its central line location $C_{CL}$, and height, $H_{CL}$ . 
(3) We define a  preliminary detected bounding box as ($C_{CL}$,$H_{CL}\times2$).
(4) Similarly, we compute the MAR from the binarized TL-Heatmap, and compute its top and bottom sides, $T_{top}$ and $T_{bottom}$.
(5) We refine the preliminary detected bounding box with $T_{top}$ and $T_{bottom}$, and generate final bounding box, as shown in Fig. \ref{fig:pipeline} (f). More results of the fine text network are presented in Fig. \ref{fig:FTN_examples}.

%% file: experiment.tex
\section{Experiment and Results}

The proposed CCTN is evaluated on three benchmarks for text localization in natural image: ICDAR 2011 \cite{Shahab2011}, ICDAR 2013 \cite{Karatzas2013} and MSRA-TD500 \cite{Yao2012}.

\subsection{Datasets and Experimental Setting}

\textbf{Benchmarks}. The ICDAR 2011 \cite{Shahab2011} includes 229 and 255 images for training and testing, respectively. There are 299 training images and 233 test images in the ICDAR 2013 \cite{Karatzas2013}. The images in both datasets are varied from 307$\times$93 to 1280$\times$960. The MSRA-TD500 database \cite{Yao2012} has 500 images containing multi-orientation text lines in different languages (e.g. Chinese, English or mixture of both). The training set contains 300 images, and the rest is used for testing. 
We follow standard evaluation protocol of the ICDAR 2011 \cite{Shahab2011,Wolf2006} and ICDAR 2013 \cite{Karatzas2013}. Evaluation on the MSRA-TD500 was originally proposed by \cite{Yao2012}, which are measured by minimum area bounding boxes.

\textbf{Training Data}.  Our training samples were generated from the training sets of the ICDAR 2011 and USTB \cite{Yin2015} databases. We randomly cropped image patches with a fixed size of 500$\times$500 from the training images. Then to increase the number of the training samples and their diversities, the cropped patches were further implemented with multiple data augmentations by rotating, flipping, and adding random noise. We finally generate 6611 training samples in total. Notice that our training samples only include English text, some of which are oriented (from the USTB dataset). We used these samples to train coarse text network. For fine text networks, we further cropped square text regions from the 6611 training samples, and enlarged them to 500$\times$500.
 

We used the pre-trained 16-layer VggNet \cite{Simonyan2015} as initialization of our networks. The newly-developed rectangle convolutional layers are initialized with a Gaussian distribution of mean 0 and standard deviation 0.01 in the training process. The learning rate and  momentum are set to $10^{-10}$ and 0.99 respectively. The coarse text network was trained with 200K iterations, while the fine text network was fine-turned on the coarse text networks with a further 100K iterations. In the test processing, each input image is resized into two scales. One scale is 500$\times$500. The other is to fix the long side to 500, while keeping shape ratio unchanged. Therefore, in our cascaded model, we use both scales for the coarse detection, and use a single-scale 500$\times$500 for the fine network, due to our square region cropping scheme.  

%

%
\subsection{Evaluation on component text networks}
\begin{table}[tb]
\centering \caption{Evaluation on individual Coarse Text Network or Fine Text Network on the ICDAR 2011.}
\begin{tabular}{|l|c|c|c|c|}

\hline
				
               & $P$             &$R$              &$F$             &$Time$ \\\hline \hline

Coarse Text Network   &0.62    &0.43       &0.50     & 0.5s \\\hline
Fine Text Network  &0.70    &0.51       &0.60    &0.3s  \\\hline
\textbf{CCTN}  &0.88    &0.79       &0.84    &1.3s\\\hline
\end{tabular}\label{tab:components}
\end{table}

\begin{figure*}
\centering
\includegraphics[height=9cm,width=16cm]{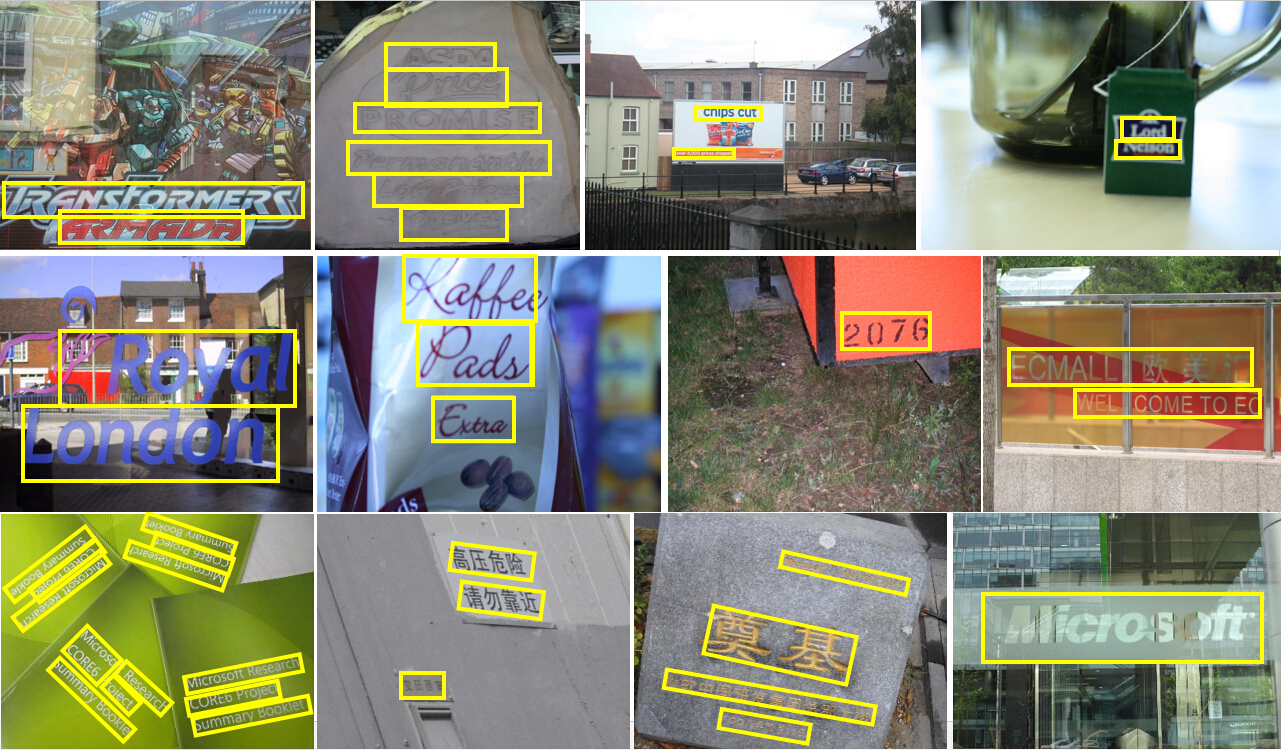}
   \caption{Full detection results on changeling examples.}
\label{fig:finalresult}
\end{figure*}

We investigate the performance of individual convolutional text networks without the cascaded architecture. The experiments were conducted on the ICDAR 2011 dataset by using each individual model for text detection. The results, including running time of each component, are reported in Table \ref{tab:components}, where $R$, $P$ and $F$ indicate Recall, Precision and F-measure, respectively. All models were implemented in Caffe framework\cite{Jia2014} with Matlab by using a single GPU.

As can be found, each individual model can not obtain reasonable performance. Obviously, our coarse-to-fine strategy significantly improves the performance, indicating that our cascaded design is indeed beneficial. It is difficult for each individual model to find accurate locations of the text lines, especially for those small-scale or multiple closing text lines.  The average running time of the individual model on a  single-scale image (500$\times$500)  is about $0.3s$. The total running time for the CCTN is about $1.3s$, including $0.5s$ for the two-scale coarse network, and $0.8s$ for the single-scale fine network. The time by the  fine network is increased due to multiple cropped regions generated from an image. The coarse network detect totally 537 text regions from all 255 text images, with \emph{only 35 false detections in all  images} (We consider an false detection as a region that dose not include any text information), compared to \emph{1000$\sim$2000 per image} error detections reported in \cite{Yin2014, Zhang2015}.  This suggests that our detector is extremely robust against background noise and outliers by considering a whole text region as a detection object, as shown in the heat-maps in Fig. \ref{fig:heatmap_ctn}. These false detections can be further removed by the fine text networks, as shown in Fig. \ref{fig:pipeline}. 

\subsection{Evaluation on full text detection}
 The full text detection results on a number of challenging images are presented in Fig.~\ref{fig:finalresult}. The results demonstrate high performance of our CCTN, which has powerful discriminative ability to detect extremely ambiguous text lines, with strong robustness against multiple text variations and significantly cluttered background.
 
We evaluate the CCTN on two benchmarks: ICDAR 2011 and ICDAR 2013.  The performance is compared extensively with most recent results in Table~\ref{tab:results_ICDAR2011} and \ref{tab:results_ICDAR2013}. Our CCTN obtains the best results, with $0.84$ and $0.86$ F-measure on the ICDAR 2011 and 2013 datasets, surpassing all previous results compared by a large margin.  In the ICDAR 2013, it outperforms the cloest TextFlow \cite{Tian2015} substantially with a $6\%$ improvement on F-measure. The large improvement mainly comes from significantly higher recall by our detector. This can be ascribed to our text region detection scheme which has strong capability for detecting highly ambiguous and low-quality text lines by jointly considering neighboring text strings, resulting in a more reliable detection, as shown in Fig. \ref{fig:main} and \ref{fig:finalresult}.

Although Zhang \emph{et al}.'s approach also detect a group of characters, they may lost recall in the bottom-up steps which connect the detected discrete short lines into text-line candidates. Furthermore, Zhang \emph{et al}.'s detector builds on symmetry property of the text line, which is sensitive to text orientations. It require a large number of detector scales (e.g., 14 scales) to reach a good recall, which raises its computational cost substantially, e.g., resulting in about $60s$ per image. Alternatively, our top-to-down coarse-to-fine strategy provides an more efficient and accurate approach for this task.

\begin{table}[tb]
\centering \caption{Experimental results on the ICDAR 2011 dataset.}
\begin{tabular}{|l|c|c|c|c|}

\hline

Method                                                &Year    &$P$         &$R$    &$F$   \\\hline \hline
\textbf{CCTN}                              & --
&\textbf{0.88}       &\textbf{0.79}    &\textbf{0.84}  \\\hline\hline
TextFlow  \emph{et al}.\cite{Tian2015}                 & 2015     &0.86       &0.76    &0.81  \\\hline
Jaderberg  \emph{et al}.\cite{Jaderberg2015}          & 2015     &-           &-    &0.81  \\\hline
Zhang  \emph{et al}.\cite{Zhang2015}                 & 2015     &0.84       &0.76    &0.80  \\\hline
MSERs-CNN \cite{Huang2014}                           & 2014     &0.88       &0.71    &0.78  \\\hline
Yin \emph{et al}.~\cite{Yin2014}                      & 2014   &0.86       &0.68    &0.76 \\\hline
Neumann $\&$ Matas~\cite{Neumann2014}                 & 2013   &0.85        &0.68    &0.75 \\\hline
SFT-TCD~\cite{Huang2013}                              & 2013   &0.82       &0.75    &0.73 \\\hline
Shi \emph{et al}.~\cite{Shi2013}                      & 2013   &0.83       &0.63    &0.72 \\\hline
\end{tabular}\label{tab:results_ICDAR2011}
\end{table}

\begin{table}[tb]
\centering \caption{Experimental results on the ICDAR 2013 dataset. Our performance is compared to the last published results in~\cite{Tian2015}.}
\begin{tabular}{|l|c|c|c|c|}

\hline

Method                                                &Year    &$P$         &$R$    &$F$   \\\hline \hline
\textbf{CCTN}                              & --
&\textbf{0.90}       &\textbf{0.83}    &\textbf{0.86}  \\\hline\hline
TextFlow  \emph{et al}.\cite{Tian2015}                 & 2015     &0.85       &0.76    &0.80  \\\hline
Zhang  \emph{et al}.\cite{Zhang2015}                 & 2015     &0.88       &0.74    &0.80  \\\hline
Lu  \emph{et al}.\cite{Lu2015}                 & 2015     &0.89       &0.70    &0.78  \\\hline
Neumann and Matas \cite{Neumann2015b} &2015    &0.82      &0.72    & 0.77 \\\hline
FASText \cite{Busta2015} & 2015 &0.84 & 0.69 &0.77\\\hline
iwrr2014 ~\cite{Zamberletti2014}                      & 2014   &0.86       &0.70    &0.77 \\\hline
USTB TexStar~\cite{Yin2014}            & 2014   &0.88       &0.66    &0.76 \\\hline
Text Spotter~\cite{Neumann2012}                   & 2012   &0.88       &0.65    &0.75 \\\hline
\end{tabular}\label{tab:results_ICDAR2013}
\end{table}

Although we got the highest Precision (0.90), as discussed, our coarse detector generates a very small number of false positives,  so that it should be expected a higher value.  However, we observed that additional false detections may happen in the refinement stage when we implement the fine text network on the enlarged text regions.  Our detector does not have any post-processing step for removing these false detections, which reduces its precision in certain degree. In fact, most current approaches include such a post-processing step to improve their performance, such as \cite{Zhang2015,Jaderberg2015,Yin2015,Yin2014,Huang2013}.


\begin{table}[tb]
\centering \caption{Comparisons of CCTN with recent methods specifically designed for multi-language and multi-orientation text lines.}
\begin{tabular}{|l|c|c|c|c|}

\hline
               & Year             &$R$              &$P$             &$F$ \\\hline \hline
               \multicolumn{5}{|c|}{MSRA-TD500}
\\\hline

\textbf{CCTN}   & --    &\textbf{0.65}       &0.79      &\textbf{0.71} \\\hline
Yin \emph{et al}.~\cite{Yin2015}  &2015        &0.63        &\textbf{0.81}        &0.71 \\\hline
Yin \emph{et al}.~\cite{Yin2014}  &2014        &0.61        &0.71        &0.66 \\\hline
Yao \emph{et al}.~\cite{Yao2014}  &2014    &0.62        &0.64        &0.61 \\\hline
Kang \emph{et al}.~\cite{Kang2014} &2014   &0.62         &0.71        &0.66  \\\hline
Yao \emph{et al}.~\cite{Yao2012}  &2012    &0.63        &0.63        &0.60 \\\hline\hline

\multicolumn{5}{|c|}{ICDAR2011}
\\\hline

\textbf{CCTN}    & --     &\textbf{0.79}         &\textbf{0.88}     &\textbf{0.84}       \\\hline
Yin \emph{et al}.~\cite{Yin2015}     &2015  &0.66        &0.84        &0.74 \\\hline \hline 

\multicolumn{5}{|c|}{ICDAR2013}
\\\hline
\textbf{CCTN}   & --    &\textbf{0.83}       &\textbf{0.90}      &\textbf{0.86}     \\\hline
Yin \emph{et al}.~\cite{Yin2015}    &2015   &0.65        &0.84        &0.73 \\\hline 							    		
\end{tabular}\label{tab:results_MSRA-TD500}
\end{table}

We further evaluate generalization ability of our CCTN to multi-language and multi-orientation text lines on the MSRA-TD500. We train the CCTN with all English text lines without using the training set of MSRA-TD500. Surprisingly, our method generalizes well to other languages, such as Chinese, and is invariant to oriented text lines, as shown in Fig. \ref{fig:finalresult}. As can be found in Table \ref{tab:results_MSRA-TD500}, our results are compared favorably against the best performance achieved by Yin \emph{et al.} \cite{Yin2015}, which is specifically designed for multi-language/orientation text lines detection. On the other hand, as compared to their results on the standard ICDAR 2011 and 2013 (reported in \cite{Yin2015}), we got more than $10\%$ improvements on both datasets. This indicates that our approach is more principled to solve fundamental problem of scene text detection.

%
 
%
%

%% file: conclusion.tex
\section{Conclusion}
We have presented  a Cascaded  Convolutional Text Network
(CCTN) for text localization in natural image. 
We introduce a new top-to-down coarse-to-fine pipeline that casts previous character-based detection into direct text region estimation. It overcomes main limitations of previous bottom-up approaches, and achieves surprising robustness and discriminative power. We develop convolutional text network by designing multiple rectangle convolutions and multiple in-network fusions,  which customizes general convolutional networks towards our task. The proposed CCTN is able to handle multi-shape and multi-scale text robustly, and is invariant to  multi-language and multi-orientation text. It works reliably on both small-scale and large-scale text in single-scale images, making it computationally attractive with sharing convolutional computation. Extensive experimental results show that our method has achieved the state-of-the-art performance on three standard benchmarks.

%% file: CCTN_arxiv.bbl
\begin{thebibliography}{10}\itemsep=-1pt

\bibitem{Busta2015}
M.~Busta, L.~Neumann, and J.~Matas.
\newblock Fastext: Efficient unconstrained scene text detector, 2015.
\newblock In IEEE International Conference on Computer Vision (ICCV).

\bibitem{Chen2004}
X.~Chen and A.~Yuille.
\newblock Detecting and reading text in natural scenes, 2004.
\newblock In IEEE Computer Vision and Pattern Recognition (CVPR).

\bibitem{Epshtein2010}
B.~Epshtein, E.~Ofek, and Y.~Wexler.
\newblock Detecting text in natural scenes with stroke width transform, 2010.
\newblock In IEEE Computer Vision and Pattern Recognition (CVPR).

\bibitem{He2016P}
P.~He, W.~Huang, Y.~Qiao, C.~C. Loy, and X.~Tang.
\newblock Reading scene text in deep convolutional sequences, 2016.
\newblock The 30th AAAI Conference on Artificial Intelligence (AAAI-16).

\bibitem{He2016}
T.~He, W.~Huang, Y.~Qiao, and J.~Yao.
\newblock Text-attentional convolutional neural networks for scene text
  detection, 2015.
\newblock arXiv:1510.03283.

\bibitem{Huang2013}
W.~Huang, Z.~Lin, J.~Yang, and J.~Wang.
\newblock Text localization in natural images using stroke feature transform
  and text covariance descriptors, 2013.
\newblock In IEEE International Conference on Computer Vision (ICCV).

\bibitem{Huang2014}
W.~Huang, Y.~Qiao, and X.~Tang.
\newblock Robust scene text detection with convolutional neural networks
  induced mser trees, 2014.
\newblock In European Conference on Computer Vision (ECCV).

\bibitem{Jaderberg2015}
M.~Jaderberg, K.~Simonyan, A.~Vedaldi, and A.~Zisserman.
\newblock Reading text in the wild with convolutional neural networks.
\newblock {\em International Journal of Computer Vision (IJCV)}, 2015.

\bibitem{Jaderberg2014}
M.~Jaderberg, A.~Vedaldi, and A.~Zisserman.
\newblock Deep features for text spotting, 2014.
\newblock In European Conference on Computer Vision (ECCV).

\bibitem{Jia2014}
Y.~Jia, E.~Shelhamer, J.~Donahue, S.~Karayev, J.~Long, R.~Girshick,
  S.~Guadarrama, and T.~Darrell.
\newblock Caffe: Convolutional architecture for fast feature embedding, 2014.
\newblock In Proceedings of the ACM International Conference on Multimedia.

\bibitem{Kang2014}
L.~Kang, Y.~Li, and D.~Doermann.
\newblock Orientation robust text line detection in natural images, 2014.
\newblock In IEEE Computer Vision and Pattern Recognition (CVPR).

\bibitem{Karatzas2013}
D.~Karatzas, F.~Shafait, S.~Uchida, M.~Iwamura, L.~G. i~Bigorda, S.~R. Mestre,
  J.~Mas, D.~F. Mota, J.~A. Almazan, and L.~P. de~las Heras.
\newblock Icdar 2013 robust reading competition, 2013.
\newblock In International Conference on Document Analysis and Recognition
  (ICDAR).

\bibitem{Li2014}
Y.~Li, W.~Jia, C.~Shen, and A.~van~den Hengel.
\newblock Characterness: An indicator of text in the wild.
\newblock {\em IEEE Trans. Image Processing (TIP)}, 23:1666--1677, 2014.

\bibitem{Long2015}
J.~Long, E.~Shelhamer, and T.~Darrell.
\newblock Fully convolutional networks for semantic segmentation, 2015.
\newblock In IEEE Computer Vision and Pattern Recognition (CVPR).

\bibitem{Lu2015}
S.~Lu, T.~Chen, S.~Tian, J.-H. Lim, and C.-L. Tan.
\newblock Scene text extraction based on edges and support vector regression.
\newblock {\em International Journal on Document Analysis and Recognition
  (IJDAR)}, 18(2):125--135, 2015.

\bibitem{Matas2004}
J.~Matas, O.~Chum, M.~Urban, and T.~Pajdla.
\newblock Robust wide-baseline stereo from maximally stable extremal regions.
\newblock {\em Image and vision computing (IVC)}, 22:761--767, 2004.

\bibitem{Neumann2014}
L.~Neumann and J.~Matas.
\newblock On combining multiple segmentations in scene text recognition, 2013.
\newblock In International Conference on Document Analysis and Recognition
  (ICDAR).

\bibitem{Neumann2015b}
L.~Neumann and J.~Matas.
\newblock Efficient scene text localization and recognition with local
  character refinement., 2015.
\newblock In International Conference on Document Analysis and Recognition
  (ICDAR).

\bibitem{Neumann2012}
L.~Neumann and K.~Matas.
\newblock Real-time scene text localization and recognition, 2012.
\newblock In IEEE Computer Vision and Pattern Recognition (CVPR).

\bibitem{Shahab2011}
A.~Shahab, F.~Shafait, and A.~Dengel.
\newblock Icdar 2011 robust reading competition challenge 2: Reading text in
  scene images, 2011.
\newblock In International Conference on Document Analysis and Recognition
  (ICDAR).

\bibitem{Shi2013}
C.~Shi, C.~Wang, B.~Xiao, Y.~Zhang, and S.~Gao.
\newblock Scene text detection using graph model built upon maximally stable
  extremal regions.
\newblock {\em Pattern Recognition}, 34:107--116, 2013.

\bibitem{Simonyan2015}
K.~Simonyan and A.~Zisserman.
\newblock Very deep convolutional networks for large-scale image recognition,
  2015.
\newblock In International Conference on Learning Representation (ICLR).

\bibitem{Tian2015}
S.~Tian, Y.~Pan, C.~Huang, S.~Lu, K.~Yu, and C.~L. Tan.
\newblock Text flow: A unified text detection system in natural scene images,
  2015.
\newblock In IEEE International Conference on Computer Vision (ICCV).

\bibitem{Wang2011}
K.~Wang, B.~Babenko, and S.~Belongie.
\newblock End-to-end scene text recognition, 2011.
\newblock In IEEE International Conference on Computer Vision (ICCV).

\bibitem{Wang2010}
K.~Wang and S.~Belongie.
\newblock Word spotting in the wild, 2010.
\newblock In European Conference on Computer Vision (ECCV).

\bibitem{Wang2012}
T.~Wang, D.~Wu, A.~Coates, and A.~Y. Ng.
\newblock End-to-end text recognition with convolutional neural networks, 2012.
\newblock In International Conference on Pattern Recognition (ICPR).

\bibitem{Wolf2006}
C.~Wolf and J.~Jolion.
\newblock Object count / area graphs for the evaluation of object detection and
  segmentation algorithms.
\newblock {\em International Journal of Document Analysis}, 8:280--296, 2006.

\bibitem{Yao2014}
C.~Yao, X.~Bai, and W.~Liu.
\newblock A unified framework for multioriented text detection and recognition.
\newblock {\em Image Processing, IEEE Transactions on}, 23(11):4737--4749,
  2014.

\bibitem{Yao2012}
C.~Yao, X.~Bai, W.~Liu, Y.~Ma, and Z.~Tu.
\newblock Detecting texts of arbitrary orientations in natural images, 2012.
\newblock In IEEE Computer Vision and Pattern Recognition (CVPR).

\bibitem{Ye2015}
Q.~Ye and D.~Doermann.
\newblock Text detection and recognition in imagery: A survey.
\newblock {\em In IEEE Trans. Pattern Analysis and Machine Intelligence
  (TPAMI)}, 37:1480--1500, 2015.

\bibitem{Yin2015}
X.~C. Yin, W.~Y. Pei, J.~Zhang, and H.~W. Hao.
\newblock Multi-orientation scene text detection with adaptive clustering.
\newblock {\em IEEE Trans. Pattern Analysis and Machine Intelligence (TPAMI)},
  37:1930--1937, 2015.

\bibitem{Yin2014}
X.~C. Yin, X.~Yin, K.~Huang, and H.~W. Hao.
\newblock Robust text detection in natural scene images.
\newblock {\em IEEE Trans. Pattern Analysis and Machine Intelligence (TPAMI)},
  36:970--983, 2014.

\bibitem{Zamberletti2014}
A.~Zamberletti, L.~Noce, and I.~Gallo.
\newblock Text localization based on fast feature pyramids and multi-resolution
  maximally stable extremal regions, 2014.
\newblock In Workshop of Asian Conference on Computer Vision (ACCV).

\bibitem{Zhang2015}
Z.~Zhang, W.~Shen, C.~Yao, and X.~Bai.
\newblock Symmetry-based text line detection in natural scenes, 2015.
\newblock In IEEE Computer Vision and Pattern Recognition (CVPR).

\bibitem{Zitnick2014}
C.~L. Zitnick and P.~Doll$\acute{a}$r.
\newblock Edge boxes: Locating object proposals from edges, 2014.
\newblock In European Conference on Computer Vision (ECCV).

\end{thebibliography}
